\documentclass[journal]{IEEEtran}

\ifCLASSINFOpdf
\else
   \usepackage[dvips]{graphicx}
\fi
\usepackage{url}

\usepackage{graphicx}
\usepackage{algorithm,algorithmic}
\usepackage{multirow}
\usepackage{kotex}
\usepackage{booktabs}
\usepackage{cite}
\usepackage{amsmath,amssymb,amsfonts}
\usepackage{algorithmic}
\usepackage{graphicx,color}
\usepackage{textcomp}
\usepackage{xcolor}
\usepackage{hyperref}
\usepackage{algorithm,algorithmic}
\usepackage{multirow}
\usepackage{kotex}
\usepackage{booktabs}

\begin{document}

\title{Elderly-Contextual Data Augmentation via Speech Synthesis for 
Elderly ASR}
\IEEEspecialpapernotice{This work has been submitted to the IEEE for possible publication. Copyright may be transferred without notice, after which this version may no longer be accessible.}

\author{Minsik Lee, Seoi Hong, Chongmin Lee, Sieun Choi, Jian Kim, Jua Han and Jihie Kim
\thanks{This research was supported by the MSIT (Ministry of Science and ICT), Korea, under the ITRC (Information Technology Research Center) support program (IITP-2026-RS-2020-II201789), and the Artificial Intelligence Convergence Innovation Human Resources Development program (IITP-2026-RS-2023-00254592) supervised by the IITP (Institute for Information \& Communications Technology Planning \& Evaluation).}
\thanks{M. Lee, S. Hong, S. Choi, J. Kim, J. Han, and J. Kim are with the Department of Computer Science and Artificial Intelligence, Dongguk University, Seoul, South Korea (e-mail: \{minsik.lee, sieunchoi, jian.kim, juai, jihie.kim\}@dgu.ac.kr; hongchaenui@gmail.com). C. Lee is with the Department of Biostatistics, Harvard University, Boston, MA, USA (e-mail: chongminlee@hsph.harvard.edu). Corresponding author: Jihie Kim.}
}

\markboth{Journal of \LaTeX\ Class Files, Vol. 14, No. 8, August 2015}
{Shell \MakeLowercase{\textit{et al.}}: Bare Demo of IEEEtran.cls for IEEE Journals}

\maketitle

\begin{abstract}
Despite recent progress in automatic speech recognition (ASR), elderly ASR (EASR) remains challenging due to limited training data and the distinct acoustic and linguistic characteristics of elderly speech. In this work, we address data scarcity in EASR through a data augmentation pipeline that combines large language model (LLM)-based transcript paraphrasing with text-to-speech (TTS) synthesis. Given an elderly speech dataset, the LLM first generates elderly-contextual paraphrases of the original transcripts, and the TTS model then synthesizes corresponding speech using elderly reference speakers. The resulting synthetic audio-text pairs are merged with the original data to fine-tune Whisper without architectural modification. We further analyze the effects of augmentation ratio and reference-speaker composition in low-resource EASR. Experiments on English and Korean elderly speech datasets from speakers aged 70 and above show that the proposed method consistently improves performance over conventional augmentation baselines, achieving up to a 58.2\% reduction in word error rate (WER) compared with the Whisper baseline.
\end{abstract}


% \begin{abstract}
% Recent advances in Automatic Speech Recognition (ASR) research have yielded significant performance improvements. However, the progress in Elderly ASR (EASR) remains constrained due to limited training data and the distinct characteristics of elderly speech, such as dysarthria. Further research should design specialized algorithms or data augmentation techniques tailored for elderly speech. This study investigates data scarcity in EASR by combining Text-To-Speech (TTS) models with Large Language Models (LLM) to generate rich synthetic data, followed by ASR model fine-tuning. We further examine how data augmentation, emphasizing variation factors such as gender and elderly speaker characteristics, can enhance performance in low-resource EASR settings. We evaluate the proposed method through both quantitative experiments and qualitative analysis on English and Korean elderly speech datasets from speakers aged 70 and above. The experimental results demonstrate that the proposed approaches achieved a 58.2\% improvement in EASR performance compared to the Whisper baseline. 
% \end{abstract}

% \begin{IEEEkeywords}
% Data augmentation, elderly speech recognition, large language models, low-resource, text-to-speech
% \end{IEEEkeywords}

\IEEEpeerreviewmaketitle
\section{Introduction}
\begin{figure}[t]
\centerline{\includegraphics[width=0.8\columnwidth]{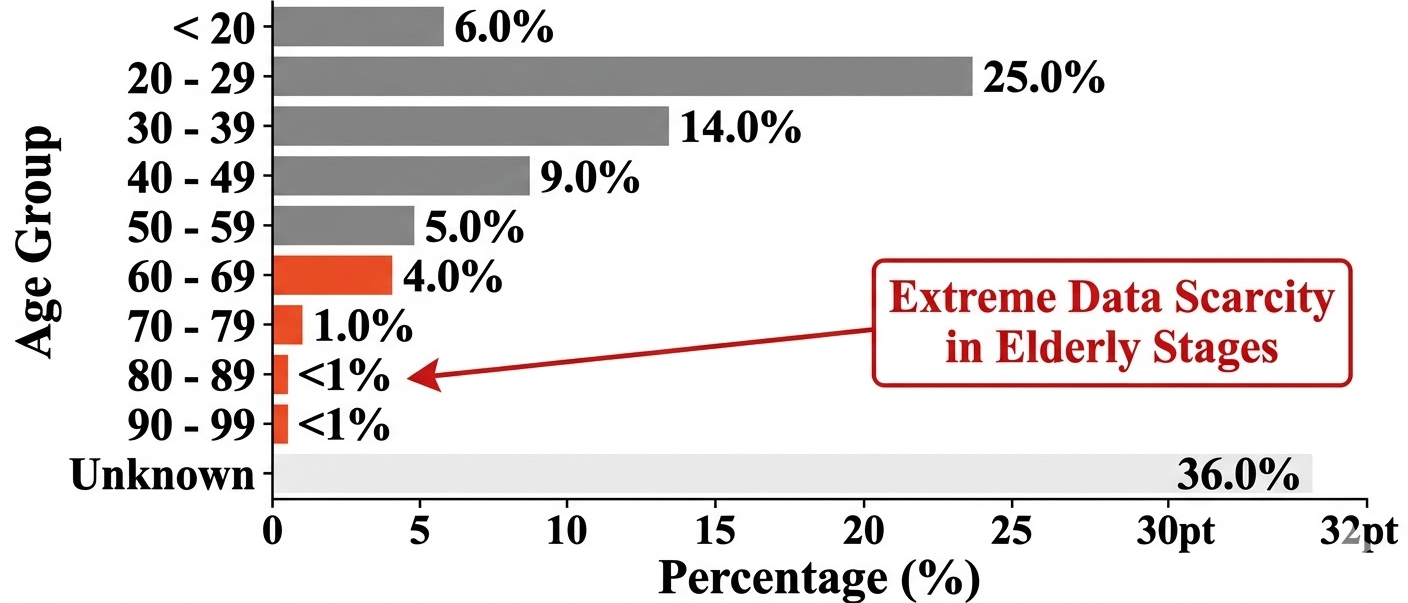}} 
    \caption{Age distribution of the CommonVoice 18.0 English corpus.}
    \label{figure_1} 
    \vspace{-5pt}
\end{figure}

\IEEEPARstart{T}{he} global elderly population is rapidly increasing, creating a growing need for speech technologies that remain reliable for older adults \cite{SCUTERI202435, chen2025seniortalk, 10.1145/3706598.3713593}. However, automatic speech recognition (ASR) systems still show clear performance degradation on elderly speech. This gap is largely caused by limited representative training data and by the distinct characteristics of elderly speech, including slower speaking rate, increased pauses, unstable articulation, hoarseness, and age-related speech disorders such as dysarthria \cite{torre2009age, miyazaki2010acoustic, winkler2003aging, pellegrini2013corpus}. A central challenge in elderly ASR (EASR) is data scarcity. Large public speech corpora, such as Common Voice \cite{ardila2019common}, contain relatively few elderly speakers in the late elderly stage, as illustrated in Fig.~\ref{figure_1}. In addition, many available elderly speech datasets are dominated by scripted or read speech and therefore do not fully reflect the conversational patterns observed in everyday elderly communication. Privacy and collection difficulty further limit the scale and accessibility of such data \cite{fukuda2023new, fukuda2020improving, sekerina2024brazilian}. As a result, prior EASR research has focused mainly on model design and training strategies \cite{ye2021development, ferri2005global, vipperla2010ageing, rudzicz2014speech, zhou2016speech, geng2022speaker, hu2023exploring, jin2023personalized, Jin_2024, 10934134}, while data augmentation for elderly speech remains underexplored.

Synthetic data generation offers a practical way to mitigate this limitation. Recent studies have shown that text-to-speech (TTS)-based augmentation can improve ASR in low-resource or high-variability settings \cite{rossenbach2020generating, bartelds2023making, leung2024training}. However, TTS-based augmentation still depends on the availability of suitable text, and elderly-relevant text is itself scarce. To address this problem, we combine large language model (LLM)-based transcript paraphrasing with TTS synthesis. The LLM first generates elderly-contextual paraphrases of existing transcripts, and the TTS model then synthesizes corresponding speech using elderly reference speakers. The resulting synthetic audio-text pairs are merged with the original data to fine-tune Whisper \cite{radford2023robust} without modifying its architecture.

Based on this framework, we investigate two questions: 1) whether LLM+TTS augmentation improves EASR beyond conventional augmentation methods, and 2) which generation settings most strongly affect its effectiveness. We analyze augmentation ratio, model size, and reference-speaker composition on English and Korean elderly speech datasets. Our contributions are as follows:
\begin{itemize}
\item We propose an EASR data augmentation framework that combines LLM-based transcript paraphrasing with TTS.

\item We demonstrate on English and Korean elderly speech datasets that the proposed method outperforms conventional augmentation baselines, with up to a 58.2\% relative reduction in in WER over Whisper.

\item We analyze generation factors, including augmentation ratio, model size, and speaker composition, and provide guidance for low-resource EASR.
\end{itemize}

\begin{figure*}[!t]
    \centerline{\includegraphics[width=\textwidth]{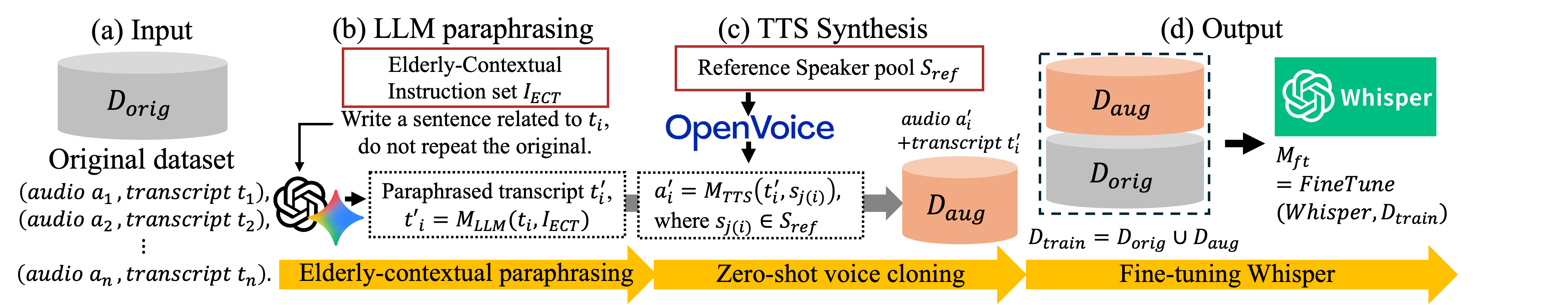}}
    \caption{Overview of the proposed EASR augmentation pipeline. An LLM first paraphrases each transcript into an elderly-contextual form, and a TTS model then generates one synthetic audio sample using a reference speaker selected from the elderly speaker pool. The resulting augmented dataset is merged with the original dataset to fine-tune Whisper.}
    \label{figure_2}
\end{figure*}
\section{Related Work}
\subsection{Acoustic and Linguistic Characteristics of Elderly Speech}
Age-related physiological changes in the laryngeal and respiratory systems lead to 
instability in acoustic measures such as jitter and shimmer \cite{suprent2025evolution}, 
which collectively make elderly speech more difficult for conventional ASR systems to 
recognize \cite{vipperla2010ageing}. Beyond acoustics, prior studies also report 
population-level linguistic tendencies, including age-related shifts in lexical 
diversity, discourse informativeness, and narrative organization 
\cite{marini2025unveiling}. These findings suggest recurrent contextual and linguistic 
tendencies at the group level rather than a fixed, uniform speaking style, motivating 
our use of elderly-contextual text generation to reflect these distributional patterns 
without assuming a single rigid style.

\subsection{Data Augmentation for Elderly ASR}
Recent work on EASR has addressed speaker-level variability through 
prompt-based adaptation of foundation models \cite{deng2025mopsa}. However, such approaches mainly increase acoustic robustness without addressing the contextual or lexical mismatch between elderly speech and general-domain training text. LLM-driven text rewriting paired with TTS synthesis has demonstrated improvements for low-resource ASR by expanding spoken-style text diversity \cite{ibaraki2025frustratingly}, yet this strategy has not been applied with elderly-specific contextual constraints.

\subsection{LLM-based Text Generation for Speech Data Augmentation}
Pairing LLM-generated transcripts with TTS synthesis has demonstrated direct gains in elderly speech recognition by improving robustness to diverse speaking styles and vocabulary \cite{rosin2025large}. Related studies further suggest that transcript-level patterns carry age-related signals, including shifts in lexical diversity and information density across the adult lifespan \cite{petriglia2025assessing}, 
motivating elderly-contextual paraphrasing as a practical source of synthetic training  data for EASR.

\section{Approach}
\subsection{Overview}
Fig.~\ref{figure_2} presents the overall pipeline of the proposed data augmentation framework. Let
$D_{orig}=\{(a_i,t_i)\}_{i=1}^{N}$ denote the original dataset, where $a_i$ is an audio
sample, $t_i$ is its transcript, and $N$ is the number of training samples. For each
transcript $t_i$, we first generate an elderly-contextual paraphrase using an LLM.
The generated transcript is then converted into synthetic speech using a TTS model
conditioned on a curated elderly reference speaker. This process yields a synthetic
audio-text pair for each original sample, and the resulting augmented dataset is merged
with the original dataset to construct the training set used for ASR model fine-tuning.
The final output of the pipeline is a fine-tuned ASR, which takes speech
as input and produces transcriptions during inference.

\subsection{Elderly-Contextual Text Generation}
To incorporate elderly-relevant linguistic characteristics, we paraphrase each transcript
into elderly-contextual text (ECT) using an LLM. 
% In the main experimental setting, we use GPT-5 as the LLM for elderly-contextual paraphrase generation. Additional comparisons with GPT-4o and Gemini 3 Flash are presented in Table~\ref{tab:quantity_analysis}-(3). 
Let $M_{\mathrm{LLM}}$ denote the LLM and
$I_{ECT}$ the instruction set designed for ECT generation. The paraphrasing process is
defined as
\begin{equation}
t'_i = M_{\mathrm{LLM}}(t_i, I_{ECT}),
\end{equation}
where $t'_i$ is the ECT paraphrase generated from the original transcript $t_i$. The
detailed prompts are summarized in Table~\ref{tab:llm_instructions}.
\begin{table}[t]
\centering
\caption{Prompt structure for LLM-based elderly-contextual paraphrasing, including formatting, paraphrasing constraints, and contextual transformation.}
\label{tab:llm_instructions}
\resizebox{\columnwidth}{!}{%
\begin{tabular}{l p{0.7\columnwidth}}
\toprule
\textbf{Step} & \multicolumn{1}{c}{\textbf{LLM Instructions and Description}} \\
\midrule
\textbf{1. Preparation} & We first present the instruction, \textit{"Make a sentence into a CSV file,"} allowing the LLM to generate sentences related to the elderly based on the transcription data. \\
\addlinespace
\textbf{2. Sentence Paraphrasing} & Specific paraphrasing guidelines are provided, such as \textit{“Write a sentence, related to the CSV file”} and \textit{“Do not repeat the original sentence.”} A word limit is also set: \textit{“Use a minimum of 3 words and a maximum of 20 words.”} \\
\addlinespace
\textbf{3. Elderly-Contextual Paraphrasing} & The final instruction guides the LLM to adjust the tone and content. For instance: \textit{“Change the given sentence into a sentence frequently used by elderly people.”} \\
\bottomrule
\end{tabular}%
}
\end{table}
\subsection{Speech Generation and Augmented Dataset Construction}
Given the generated transcript $t'_i$, we synthesize one corresponding audio sample
using a TTS model. In our experiments, we use OpenVoice2 \cite{qin2023openvoice} as
the TTS model. Let $M_{\mathrm{TTS}}$ denote the TTS model and
$S_{ref}=\{s_1,s_2,\dots,s_K\}$ denote the curated pool of $K$ elderly reference speakers.
The synthesis process is formulated as
\begin{equation}
a'_i = M_{\mathrm{TTS}}(t'_i, s_{j(i)}), \quad s_{j(i)} \in S_{ref},
\end{equation}
where $a'_i$ is the synthesized speech for transcript $t'_i$, and $s_{j(i)}$ is the
reference speaker assigned to the $i$-th generated transcript. The indexing function
$j(i)$ determines the assigned speaker, and assignments are balanced across $S_{ref}$
so that each speaker contributes approximately the same number of synthesized samples.

The generated synthetic audio-text pairs are collected into the augmented dataset
\begin{equation}
D_{aug}=\{(a'_i,t'_i)\}_{i=1}^{N},
\end{equation}
which is then merged with the original dataset to form the training dataset
\begin{equation}
D_{train}=D_{orig}\cup D_{aug}.
\end{equation}

\subsection{ASR Fine-Tuning}
The merged dataset $D_{train}$ is used to fine-tune an ASR model. In our experiments,
we use Whisper \cite{radford2023robust} as the backbone ASR model:
\begin{equation}
M_{ft}=\mathrm{FineTune}(\mathrm{Whisper}, D_{train}),
\end{equation}
where $M_{ft}$ denotes the resulting fine-tuned ASR model. During inference,
$M_{ft}$ takes speech as input and outputs transcriptions. Following prior work
\cite{suh2024improving, leung2024training}, we use Whisper without modifying its
architecture and improve EASR performance through data augmentation.

\begin{table}[t]
\centering
\caption{Performances according to the proposed method combined with SpecAugment on CV18 and VOTE400.}
\label{tab:llm_text_effect}
\resizebox{\columnwidth}{!}{%
\begin{tabular}{l cc cc}
\toprule
\multirow{2}{*}{\textbf{Models}} & \multicolumn{2}{c}{\textbf{CV18 (\%)}} & \multicolumn{2}{c}{\textbf{VOTE400 (\%)}} \\
\cmidrule(lr){2-3} \cmidrule(lr){4-5}
& \textbf{WER} & \textbf{CER} & \textbf{WER} & \textbf{CER} \\
\midrule
Baseline & 4.1 & 1.8 & 11.6 & 5.5 \\
+ TTS \cite{bartelds2023making} (W/o LLM) & 3.1 & 1.4 & 8.8 & 4.5 \\
+ TTS \cite{bartelds2023making} (W/ LLM) & 2.7 & 1.2 & 6.5 & 3.5 \\
\textbf{Proposed (LLM+TTS, ECT)} & \textbf{2.2} & \textbf{1.0} & \textbf{5.2} & \textbf{2.9} \\
% \textbf{Proposed + SpecAugment} & \textbf{2.1} & \textbf{0.9} & \textbf{4.8} & \textbf{2.7} \\
\bottomrule
\end{tabular}%
}
\end{table}

\begin{table}[t!]
\centering
\caption{Ablation of augmentation ratio using ECT-based synthetic data across Whisper model sizes.}
\label{tab:quantity_analysis}
\resizebox{\columnwidth}{!}{% 
\begin{tabular}{l c c c c c c}
\toprule
\multirow{2}{*}{\textbf{Dataset}} & \multirow{2}{*}{\textbf{Aug \%}} & \multirow{2}{*}{\textbf{Spk.}} & \multicolumn{4}{c}{\textbf{Whisper Model Performance (WER / CER)}} \\
\cmidrule(lr){4-7}
& & & \textbf{Small} & \textbf{Medium} & \textbf{Large-v2} & \textbf{Large-v3} \\
\midrule
\multirow{6}{*}{CV18} 
 & Baseline & - & 5.8 / 2.6 & 4.3 / 1.9 & 4.2 / 1.8 & 4.1 / 1.8 \\
 & +10\% & 2 & 5.4 / 2.4 & 4.0 / 1.8 & 3.8 / 1.7 & 3.8 / 1.6 \\
 & +30\% & 4 & 4.8 / 2.1 & 3.4 / 1.5 & 3.2 / 1.4 & 3.1 / 1.4 \\
 & +50\% & 8 & 3.9 / 1.7 & 2.7 / 1.2 & 2.5 / 1.1 & 2.5 / 1.1 \\
 & +70\% & 8 & 3.7 / 1.6 & 2.5 / 1.1 & 2.4 / 1.0 & 2.3 / 1.0 \\
 & \textbf{+100\%} & \textbf{8} & \textbf{3.6 / 1.5} & \textbf{2.4 / 1.0} & \textbf{2.3 / 0.9} & \textbf{2.1 / 0.9} \\
\midrule
\multirow{6}{*}{VOTE400} 
 & Baseline & - & 26.5 / 12.5 & 17.5 / 8.3 & 12.8 / 6.1 & 11.6 / 5.5 \\
 & +10\% & 2 & 25.2 / 12.0 & 16.8 / 8.0 & 12.2 / 5.8 & 10.9 / 5.2 \\
 & +30\% & 4 & 21.8 / 10.5 & 13.5 / 6.5 & 9.4 / 4.5 & 8.4 / 4.0 \\
 & +50\% & 8 & 19.5 / 9.4 & 10.8 / 5.2 & 9.6 / 4.6 & 9.4 / 4.5 \\
 & +70\% & 8 & 19.0 / 9.2 & 10.4 / 5.0 & 8.0 / 3.9 & 7.6 / 3.7 \\
 & \textbf{+100\%} & \textbf{8} & \textbf{18.2 / 8.9} & \textbf{10.1 / 4.9} & \textbf{6.5 / 3.2} & \textbf{4.8 / 2.7} \\
\bottomrule
\end{tabular}%
}
\end{table}

\section{Experimental Setup}

\subsection{Datasets}
To evaluate the proposed augmentation framework for EASR, we used one English dataset and one Korean dataset. Unlike prior studies that often define elderly speakers as those aged 65 and above \cite{werner2011older, orimo2006reviewing}, we focus on the late elderly group, specifically speakers aged 70 years and older.

\par
\noindent
{\bf Common Voice 18.0 (English):}
We used the English subset of Common Voice 18.0 (CV18) \cite{ardila2019common}, a large-scale open-source ASR corpus. From this dataset, we extracted 7,480 utterances from speakers aged 70 and above, covering 12 regional accents. The resulting subset contains 12 hours and 57 minutes of speech, with 77\% male and 23\% female speakers. We used 5,984 utterances for training, 1,197 for validation, and 299 for testing.

\par
\noindent
{\bf VOTE400 (Korean):}
We used VOTE400 (Voice of the Elderly 400 Hours) \cite{jang2021vote400}, a Korean speech dataset designed for elderly-care voice interaction research. The dataset consists of speech recordings from 104 speakers aged 75 and above, collected through regional senior welfare institutes. It includes speakers from five regions of South Korea, reflecting regional accent and dialect variation. For our experiments, we used 100,000 utterances (10 hours in total), with 80,000 for training, 16,000 for validation, and 4,000 for testing.

For evaluation, we report both Word Error Rate (WER) and Character Error Rate (CER) for both datasets. Following ASR practice, WER is used as the primary metric for the English CV18 dataset, while CER is used as the primary metric for the Korean VOTE400 dataset.

\subsection{Implementation Details}
We used the pre-trained Whisper ASR model \cite{radford2023robust} as the baseline and fine-tuned it for our experiments. All audio data were resampled to 16 kHz. Training was conducted for 5 epochs using the AdamW optimizer \cite{loshchilov2017decoupled} with a warm-up scheduler, a learning rate of 1e–05, and a weight decay of 0.01. The training process employed a batch size of 16. All models were trained on an NVIDIA RTX 4090, requiring approximately 6 hours for CV18, 7  hours for VOTE400. 

For the main experiments, GPT-5 was used for ECT generation. We additionally evaluated GPT-4o and Gemini 3 Flash in the LLM comparison reported in Table~\ref{tab:comprehensive_analysis}-(3). To ensure reproducibility, we used the following generation settings: GPT-4o and GPT-5 (temperature=0, top p=0.9, max tokens=512, frequency penalty=0.2, presence penalty=0.1), and Gemini 3 Flash (temperature=1.0, thinking level=‘low’).
To evaluate the statistical significance of the differences observed in our results, we employed the Wilcoxon signed-rank test\cite{wilcoxon1945individual}. Unless otherwise stated, all statistical tests were conducted at a significance level of p $<$ 0.05.

\section{Experimental Results}
\subsection{Main Results}
Unless otherwise noted, all ECT-based results use GPT-5 for paraphrase generation. Table~\ref{tab:llm_text_effect} compares TTS-only augmentation, standard LLM-based text generation without any specific prompt design, and the proposed ECT-based setting. TTS-only augmentation already improves over the baseline, showing the benefit of synthetic speech. Adding LLM-generated text provides further gains, and the ECT-based setting achieves the best performance on both datasets.
% , yielding up to a 58.2% relative error reduction when further combined with SpecAugment.
% 0323 Proposed + SpecAugment 내용을 추가해야 할까요? 저희 Main Results 작성할때 TABLE II에  Proposed + SpecAugment 항목이 있는데 뺴는게 좋을까요?  C. Comparison of Data Augmentation에 TABLE IV에 Proposed + SpecAugment 에 있어야할지 지금은 둘다 있는데 Main Results에 있으면 C 마지막문단을 수정해야하지 않을까 싶어서요. 아니면 둘다 있어도 좋아 보인다면 수정하지 않을 생각입니다. 

% These results indicate that elderly-contextual text generation contributes beyond TTS alone.
% generic LLM-based text generation, 보다 리뷰 내용을 반영해서 3. Reason for performance improvement - The experiments though quite rich, fail to highlight that LLM paraphrasing would have an important role in the methodology. If we just use elderly speakers to condition the OpenVoice 2 TTS model without using LLM paraphrasing, what would be the impact? 
% standard LLM-based text generation without any specific prompt design, 으로 수정함.  

% 0323 수정 버전 확인 부탁드립니다. 
\subsection{Quantitative Analysis on CV18.0 and VOTE400}
Table~\ref{tab:quantity_analysis} presents the augmentation ratio using ECT synthetic data across different Whisper model sizes. Overall, increasing the amount of ECT-based augmented data improved performance on both CV18 and VOTE400, with the best results obtained at 100\% augmentation. While scaling up the model size naturally yielded better baseline recognition, the proposed ECT augmentation provided substantial additional reductions in the primary error metrics WER and CER for model. Interestingly, the augmentation proved highly effective for smaller models enabling the augmented Whisper-small to surpass the baseline performance of much larger models on CV18 while still delivering improvements for the most capable Whisper-large-v3. These results show that ECT augmentation is consistently effective across model sizes and that increasing the augmentation ratio is beneficial for EASR.

\begin{table}[t]
\centering
\caption{Results of Data Augmentation Methods, Speaker Balance, and LLM on EASR Performance.}
\label{tab:comprehensive_analysis}
\resizebox{\columnwidth}{!}{%
\begin{tabular}{l c c}
\toprule
\textbf{Method / Factor} & \textbf{CV18 (WER / CER)} & \textbf{VOTE400 (WER / CER)} \\
\midrule
\multicolumn{3}{r}{\textit{(1) Comparison of Data Augmentation Methods}} \\
\midrule
Baseline & 4.1 / 1.8 & 11.6 / 5.5 \\
+ Speed perturbation \cite{ko15_interspeech} & 3.9 / 1.7 & 10.7 / 5.1 \\
+ SpecAugment \cite{park2019specaugment} & 3.6 / 1.5 & 10.1 / 4.8 \\
Proposed (LLM+TTS, ECT) & 2.2 / 1.0 & 6.1 / 2.9 \\
Proposed + Speed perturbation & 2.1 / 0.9 & 5.9 / 2.9 \\
\textbf{Proposed + SpecAugment} & \textbf{2.1 / 0.9} & \textbf{4.8 / 2.7} \\
\midrule
\multicolumn{3}{r}{\textit{(2) Effect of Gender Balance}} \\
\midrule
8F + 0M (Female Only) & 3.4 / 1.5 & 8.2 / 3.9 \\
6F + 2M  & 2.9 / 1.3 & 7.5 / 3.6 \\
\textbf{4F + 4M (Balanced)} & \textbf{2.7 / 1.2} & \textbf{6.9 / 3.3} \\
2F + 6M  & 2.8 / 1.2 & 7.3 / 3.5 \\
0F + 8M (Male Only) & 3.2 / 1.4 & 7.8 / 3.7 \\
\midrule
\multicolumn{3}{r}{\textit{(3) Effect of LLM (ECT Paraphrasing)}} \\
\midrule
Gemini 3 Flash & 2.3 / 1.0 & 6.5 / 3.1 \\
GPT-4o & 2.2 / 1.0 & 5.8 / 2.8 \\
\textbf{GPT-5} & \textbf{2.1 / 0.9} & \textbf{4.8 / 2.7} \\
\bottomrule
\end{tabular}%
}
\end{table}

\begin{table}[t!]
\centering
\caption{Qualitative Analysis Examples, (A) Examples of ASR Outputs, Highlighting Corrections for Phonetic Ambiguity in English (CV18) and Domain-Specific Lexicon in Korean (VOTE400). (B) Examples of Elderly-Contextual Paraphrasing From CV18.}
\label{tab:qualitative_combined_revised}

% Part A: ASR Transcription Examples
\resizebox{\columnwidth}{!}{%
\begin{tabular}{c p{0.54\textwidth}}
\toprule
\multicolumn{2}{c}{\textbf{Part A: ASR Transcription Correction Examples}} \\
\midrule
\addlinespace
\multicolumn{2}{l}{\textbf{CV18 (English Example)}} \\
\midrule
\parbox{2.5cm}{\centering \textit{Reference}} & My \textbf{\textcolor{blue}{grandchild}} has grown up so much. \\
\parbox{2.5cm}{\centering \textbf{\textit{Proposed}}} & My \textbf{\textcolor{blue}{grandchild}} has grown up so much. \textit{(Correct)} \\
\parbox{2.5cm}{\centering \textit{Baseline}} & My \textbf{\textcolor{red}{and child}} has grown up so much. \textit{(Incorrect)} \\
\midrule
\addlinespace
\multicolumn{2}{l}{\textbf{VOTE400 (Korean Example)}} \\
\midrule
\parbox{2.5cm}{\centering \textit{Reference}} & 오늘 병원에 가서 \textbf{\textcolor{blue}{혈압 약(hyeorap yak)}} 타왔어. \\
\parbox{2.5cm}{\centering \textbf{\textit{Proposed}}} & 오늘 병원에 가서 \textbf{\textcolor{blue}{혈압 약(hyeorap yak)}} 타왔어. \textit{(Correct)} \\
\parbox{2.5cm}{\centering \textit{Baseline}} & 오늘 병원에 가서 \textbf{\textcolor{red}{결합 약(gyeorap yak)}} 타왔어. \textit{(Incorrect)} \\
\bottomrule
\end{tabular}%
}

\vspace{2mm} % Add some vertical space between the two parts

% Part B: LLM Paraphrasing Style Examples
\resizebox{\columnwidth}{!}{%
\begin{tabular}{c p{0.54\textwidth}}
\toprule
\multicolumn{2}{c}{\textbf{Part B: LLM Paraphrasing Examples}} \\
\midrule
\parbox{2.5cm}{\centering \textit{Reference}} & We speak of them only to children. \\
\midrule
\parbox{2.5cm}{\centering Retrospective} & \textbf{These old stories} are only ever spoken of to the grandchildren, just like always. \\
\midrule
\parbox{2.5cm}{\centering Relational} & Only to \textbf{our dear grandchildren} are these special kinds of stories ever told. \\
\midrule
\parbox{2.5cm}{\centering Opinion} & \textbf{I believe} some stories should only be passed down directly to one's grandchildren. \\
\midrule
\parbox{2.5cm}{\centering Didactic} & \textbf{You see,} it is a great responsibility to pass these stories to grandchildren. \\
\midrule
\parbox{2.5cm}{\centering Worry/Concern} & \textbf{I just worry} the grandchildren won't listen to these stories we tell them. \\
\midrule
\parbox{2.5cm}{\centering Philosophical} & \textbf{In the end, our stories only live on through our beloved grandchildren.} \\
\midrule
\parbox{2.5cm}{\centering Humorous} & We only tell these stories to grandchildren; \textbf{they’re too old for anyone else!} \\
\midrule
\parbox{2.5cm}{\centering Agreeing} & \textbf{Perhaps it's best} that we only speak of these matters to the grandchildren. \\
\midrule
\parbox{2.5cm}{\centering Complaining} & We speak of them only to the grandchildren; \textbf{nobody else seems to care anymore.} \\
\midrule
\parbox{2.5cm}{\centering Inquisitive} & \textbf{I wonder why} we are only supposed to speak of these things to grandchildren? \\
\bottomrule
 % End of the second resizebox
\end{tabular}%
}
\end{table}

\subsection{Comparison of Augmentation, Speaker Composition, and LLM}
\label{secV-C}
Table~\ref{tab:comprehensive_analysis} summarizes three additional analyses: data augmentation methods, reference-speaker composition, and LLM for ECT paraphrasing. First, among the augmentation methods in Table~\ref{tab:comprehensive_analysis}-(1), the proposed LLM+TTS framework substantially outperformed conventional signal-level augmentation methods such as speed perturbation and SpecAugment. Further combining the proposed method with SpecAugment yielded the best overall performance, achieving 2.1 on CV18 and 2.7 on VOTE400. This result suggests that elderly-contextual text generation and signal-level augmentation are complementary.

Second, Table~\ref{tab:comprehensive_analysis}-(2) shows that reference-speaker composition also affects performance. The balanced setting (4F+4M) achieved the best results on both datasets, indicating that a balanced speaker pool produces more effective synthetic training data than strongly skewed gender compositions.

Finally, Table~\ref{tab:comprehensive_analysis}-(3) compares different LLMs for ECT paraphrasing. Although the performance differences are modest, GPT-5 achieved the best results, followed by GPT-4o and Gemini 3 Flash. This trend suggests that LLM choice may affect the quality of generated elderly-contextual paraphrases and the resulting EASR performance.

\subsection{Qualitative Analysis}

The qualitative examples in Table~\ref{tab:qualitative_combined_revised} further support the effectiveness of the proposed approach. As shown in Part A, our method corrects common baseline errors caused by acoustic ambiguity (e.g., ``grandchild'' vs. ``and child'') and domain-specific vocabulary (e.g., ``hyeorap yak''). This suggests that the proposed augmentation improves both acoustic robustness and domain relevance. Part B shows that the LLM can generate diverse paraphrases from a single neutral sentence by introducing elderly-contextual variation. At the same time, some outputs remain stereotypical or formulaic. For example, the model occasionally prefixes sentences with expressions such as ``Back in my day \ldots'' and tends to replace words such as ``child'' with more archetypal alternatives like ``grandchildren,'' even when other plausible variations are possible. Overall, the improved transcription accuracy and increased contextual diversity help explain the performance gains observed in our experiments.

\section{Conclusion}
We proposed a data augmentation framework for EASR that combines LLM-based elderly-contextual paraphrasing with TTS synthesis. The generated synthetic audio-text pairs improved Whisper-based EASR on both English and Korean datasets and outperformed conventional augmentation baselines. We further showed that augmentation ratio and reference-speaker composition are important factors in constructing effective synthetic training data. This work is limited by its evaluation on only two languages and by its dependence on prompt design, reference-speaker selection, and current TTS quality. Future work will expand the evaluation to more languages and speaker groups, improve control over LLM-generated text quality, and explore closer integration between generative models and ASR training.

\end{document}